\documentclass{article}

\usepackage[preprint]{corl_2026} %

\title{\methodname: Human-in-the-Loop Adaptation of Generative Robot Policies in Latent Space}

\usepackage{authblk}
\usepackage{graphicx}
\usepackage{amsmath}
\usepackage{amssymb}
\usepackage{xspace}
\usepackage{booktabs}
\usepackage{xcolor}
\usepackage{wrapfig}
\usepackage{cleveref}
\crefname{appendix}{Appendix}{Appendices}
\Crefname{appendix}{Appendix}{Appendices}
\creflabelformat{equation}{#2#1#3} %

\newcommand{\rwimg}[2][0.17\textwidth]{%
  {\setlength{\fboxsep}{0pt}%
   \fcolorbox{black!40}{white}{%
     \parbox[c][0.9in][c]{#1}{\centering%
       \includegraphics[width=#1,height=0.9in]{#2}%
     }%
   }}%
}

\newcommand{\methodname}{FlowDAgger\xspace}

\author[1]{Michael Murray}
\author[1,4]{Daphne Chen}
\author[1]{Simran Bagaria}
\author[1]{Dean Fortier}
\author[1]{Tess Hellebrekers}
\author[1]{Galen Mullins}
\author[1]{Harshavardhan Gajarla}

\author[2,3]{Oier Mees}
\author[4]{Maya Cakmak}
\author[1]{Andrey Kolobov}
\affil[1]{Microsoft Research}
\affil[2] {Microsoft}
\affil[3] {ETH Zurich}
\affil[4]{University of Washington}

\begin{document}
\maketitle

\begin{abstract}
    Pretrained generative robot policies based on flow matching and diffusion have achieved impressive results across a wide range of manipulation tasks. Yet real-world deployments routinely expose failure modes outside the pretraining distribution. Closing these gaps typically requires large-scale data collection or online reinforcement learning on physical hardware, which is impractical for rapid and safe adaptation. We present \methodname, a sample- and compute-efficient method for adapting frozen generative robot policies from human interventions in latent space. Our key idea is action inversion: each human expert action is mapped to the noise that would have produced it under the frozen base policy, using reverse-time integration followed by local refinement. The resulting inverted noise provides supervision for a lightweight latent policy that steers the base model at deployment time, enabling rapid skill acquisition while preserving its behavioral priors. We evaluate \methodname in simulation and on real-world bimanual and single-arm manipulation, adapting both action-head VLAs and world-action models from a handful of interventions. \methodname outperforms supervised fine-tuning and latent-space RL baselines and preserves pretrained skills on held-out tasks, offering a practical path for adapting robot foundation models in the real world. Website: \url{https://microsoft.github.io/FlowDAgger}
\end{abstract}


\section{Introduction}
\label{sec:intro}

    Foundation models for robot manipulation have achieved impressive results on a wide range of tasks by leveraging large-scale demonstration datasets and expressive generative architectures. The dominant approach across modern systems is a learned generative process that maps random noise to actions conditioned on observations. Building on the visuomotor diffusion policy~\citep{chi2023diffusion}, this approach underlies recent vision-language-action (VLA) models~\citep{intelligence2025pi05, nvidia2025groot}, world-action models (WAM)~\citep{kim2026cosmospolicy, ye2026dreamzero,hou2026world}, and large multitask diffusion policies~\citep{tri2025lbm,dasari2024ingredients}. Trained on large, diverse demonstration corpora, these policies encode broad behavioral priors that transfer surprisingly well across embodiments and tasks. However, in any specific deployment they routinely fail due to unfamiliar objects, novel scene dynamics, embodiment quirks, and long-tail edge cases not covered by the pretraining mixture.

    Existing approaches for closing these gaps are problematic. Collecting additional demonstrations to cover the state-action space for subsequent offline finetuning is tedious; finetuning large-scale models is computationally expensive; moreover, the resulting policy tends to erode broader skills already present in the base model. Interactive imitation in the action space~\citep{ross2011dagger, kelly2019hgdagger, mandlekar2020iwr} handles the covariate-shift problem, but it, too, works by adapting the entire generative policy, an expensive and unstable update that can corrupt the learned prior; residual methods~\citep{xu2025crdagger, jiang2024transic} keep the base frozen and learn an additive correction, but apply it in the unconstrained action space, where it can push the policy off the behaviors the base reliably produces. These failures suggest that the adaptation interface matters: weight-space updates can damage the pretrained prior, while unconstrained action-space corrections can leave the support of the generative policy. Latent-space reinforcement learning, exemplified by DSRL~\citep{wagenmaker2025dsrl}, instead trains a lightweight controller in the policy's noise space. This is markedly more sample-efficient, but relies on a reward signal and autonomous exploration, both of which are bottlenecks for real-world deployment.

    In this work, we adapt generative robot policies from human corrections by learning in the policy's latent space. Prior latent-space methods show that a frozen generative policy can be steered effectively through its noise variables, but human experts intervene by providing corrective actions, not noise targets. Key to our approach is \emph{action inversion} -- a mechanism for mapping a human expert's intervention actions into corresponding latent-space vectors via reverse-time integration and local refinement. Given an observation $s$ and an expert action $a^*$, we invert $a^*$ to a noise vector $w^*$ such that the base policy's generative process from $w^*$ at $s$ reproduces $a^*$. Each $(s, w^*)$ pair becomes supervision for a small latent policy that operates within the generative process of the frozen base policy. We never modify the base policy's weights, relying only on forward passes through the base model and adapting it entirely through its input noise.

    We refer to this approach as \textbf{\methodname}, and instantiate it for adapting flow-matching and diffusion-based policies in the real world. We evaluate \methodname in simulation and on physical bimanual and single-arm manipulation, adapting both action-head VLAs and world-action models. Across these settings, \methodname reaches target success rates with substantially fewer human interventions than offline fine-tuning via behavior cloning and action-space DAgger, and with fewer environment interactions than latent-space reinforcement learning. Because the base policy is never modified, \methodname preserves pretrained skills on held-out tasks, and adaptation remains light enough to train on a consumer GPU.

\begin{figure}[t]
    \centering
    \includegraphics[width=0.85\columnwidth]{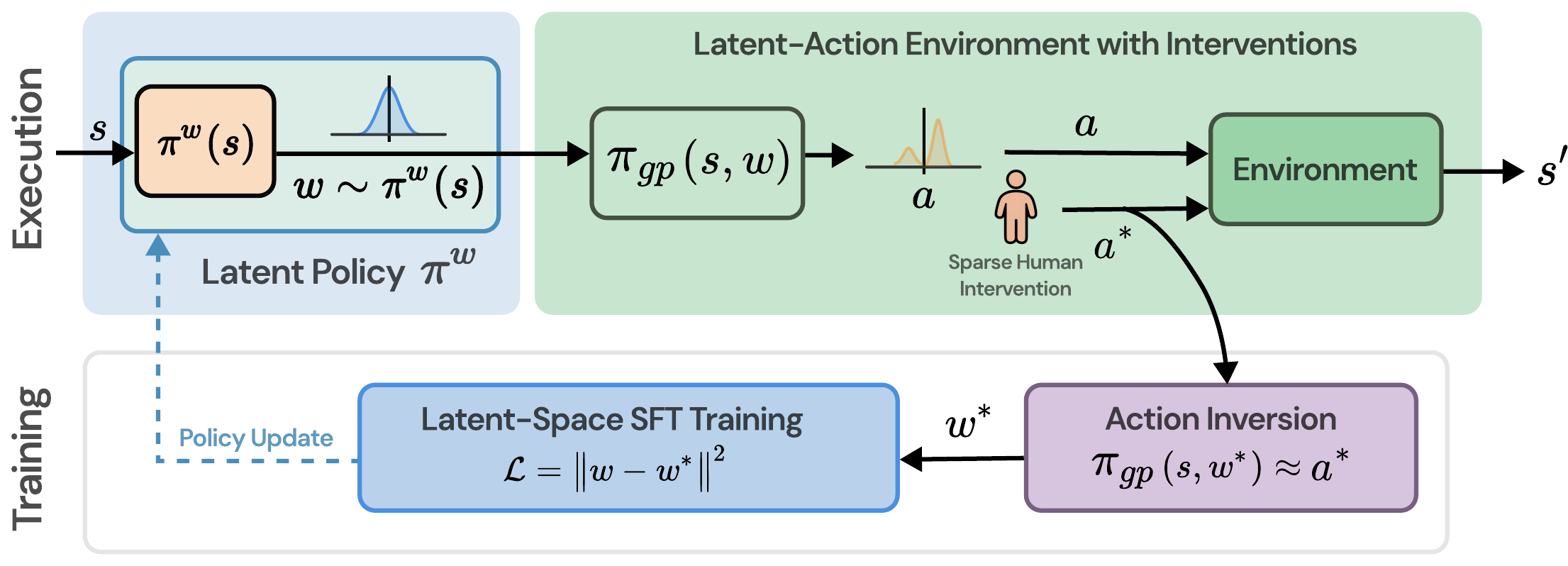}
    \caption{An overview of \textsc{\methodname}. A pre-trained generative policy $\pi_{\mathrm{gp}}$ producing actions $a$ is deployed with a human operator in the loop. When the operator intervenes, the corrective action $a^*$ is inverted back to a vector $w^*$ in the policy's latent space. The resulting $(s, w^*)$ pairs supervise a lightweight latent-space policy that adapts the policy without modifying the base model's weights.}
    \label{fig:method_overview}
\end{figure}

\section{Related Work}
\label{sec:related}
    \paragraph{Generative robot policies.}
    Diffusion Policy~\citep{chi2023diffusion} showed that diffusion-based action heads can capture multi-modal manipulation behaviors directly from demonstrations, and the approach has since been scaled to large multitask settings~\citep{tri2025lbm,dasari2024ingredients,octo_2023}. Flow matching~\citep{lipman2023flow}, a closely related approach, underlies recent VLAs~\citep{intelligence2025pi05, nvidia2025groot} and WAMs~\citep{kim2026cosmospolicy, ye2026dreamzero,pai25,hou2026world}. \methodname targets this generative subfamily: any policy whose action generation is the integration of an ODE driven by a noise sample.

    \paragraph{Interactive imitation and human-in-the-loop learning.}
    DAgger~\citep{ross2011dagger} introduced on-policy corrective supervision to address covariate shift in behavioral cloning, and HG-DAgger~\citep{kelly2019hgdagger} and IWR~\citep{mandlekar2020iwr} made human-gated variants practical for manipulation. Sirius~\citep{liu2023sirius} and Sirius-Fleet~\citep{liu2024siriusfleet} extend this paradigm with shared-autonomy deployment and trust-weighted behavior cloning, scaling human-in-the-loop learning to robot fleets. HIL-SERL~\citep{luo2024hilserl} and RLIF~\citep{luo2024rlif} fold interventions into an RL loop instead, combining human corrections with a reward signal to drive policy improvement. Building on residual policy learning~\citep{silver2018residual, johannink2018residualreinforcementlearningrobot}, residual DAgger~\citep{xu2025crdagger, jiang2024transic} keeps the base policy frozen and learns an additive action-space correction from human interventions. \methodname shares the intervention-driven data collection of these methods but supervises in the policy's noise space.

    \paragraph{Latent and noise-space policy learning.}
    DSRL~\citep{wagenmaker2025dsrl} is the closest methodological neighbor to our approach. DSRL trains a small controller in the noise space of a frozen diffusion policy using RL, and demonstrates substantial sample efficiency relative to weight-adaptation fine-tuning of the base policy. \methodname shares DSRL's architectural premise -- a lightweight policy acting in the noise input of a frozen generative base -- and differs in the source of supervision: DSRL learns from autonomous rollouts under a reward function, whereas \methodname learns from human corrections inverted back into noise space. Latent-action methods more broadly~\citep{lynch2019play} compress demonstrations into a learned action space; we instead operate in the noise space the base policy already exposes.

    \paragraph{Inversion in generative models.}
    Inverting a generated sample back to its initial noise is a well-studied operation in image synthesis. DDIM inversion~\citep{song2021ddim} recovers the deterministic latent for a diffusion sample; null-text inversion~\citep{mokady2023nulltext} improves fidelity under classifier-free guidance; and analogous techniques have been developed for rectified-flow models~\citep{rout2024rfinversion}. \methodname brings inversion-based adaptation to both flow-matching action heads and WAMs. This requires two modifications of the standard inversion machinery. For flow-matching action heads, we use a per-step fixed-point variant suited to their few-step ODE schedules, where the trajectory-level methods standard in image diffusion become unstable (\Cref{sec:method:inversion}). For WAMs, where no action-only process exists, we invert the joint world-action diffusion directly, recovering a joint-latent noise from a target in which only the action frame carries the correction (\Cref{sec:method:wam}).

\section{Background}
\label{sec:background}
\paragraph{Generative action policies.}
We consider a pretrained \emph{generative policy} $\pi_{\mathrm{gp}}: \mathcal{S} \rightarrow \mathcal{P}(\mathcal{A})$ that maps an observation $s \in \mathcal{S}$ to a distribution over an action space $\mathcal{A}$, where $a \in \mathcal{A}$ is an action chunk of horizon $H$. We use ``observation'' for the policy's full input, which may include a history of past sensor readings. Given $s$, the policy draws a noise sample $w \sim \mathcal{N}(0, I)$ from the noise space $\mathcal{W}$ and transforms it into $a$ by integrating an ordinary differential equation (ODE) defined by a learned velocity field~$v_\theta$:
\begin{equation}
    \frac{dx}{dt} = v_\theta(x, t, s), \qquad x(0) = w, \qquad a = x(1).
\end{equation}
We write $\pi_{\mathrm{gp}}(s, w)$ for the resulting noise-to-action map, $a = \pi_{\mathrm{gp}}(s, w)$, and $\pi_{\mathrm{gp}}(a \mid s)$ for the action distribution it induces when $w \sim \mathcal{N}(0, I)$. Both flow-matching~\citep{lipman2023flow}  and diffusion policies~\citep{chi2023diffusion} fit this description: the former learns the velocity field $v_\theta$ directly, while the latter's reverse process admits an equivalent probability-flow ODE. \methodname applies to any base policy of this form.

\paragraph{Euler discretization.}
In practice, the map is evaluated with a $K$-step Euler discretization of step size $\Delta t = 1/K$. With $x_0 = w$ and $t_k = k\,\Delta t$, the action is produced by the recurrence
\begin{equation}
    x_{k+1} = x_k + \Delta t \, v_\theta(x_k, t_k, s), \qquad k = 0, \dots, K-1, \qquad a = x_K.
    \label{eq:euler}
\end{equation}
Flow-matching action heads typically use a small number of steps ($K$ on the order of 10), in contrast to the long schedules ($K$ in the 100s) common in image diffusion. This few-step regime is central to the inversion procedure in \Cref{sec:method:inversion}.

\paragraph{Noise as the adaptation surface.}
For a fixed observation $s$, the base policy is a deterministic map $\pi_{\mathrm{gp}}(s, \cdot)$ from noise to action, so the noise sample $w$ alone determines the action. The action distribution is therefore controlled entirely by the distribution over $\mathcal{W}$. This exposes the noise space as a surface for adapting the policy without modifying $\theta$: replacing the standard draw $w \sim \mathcal{N}(0, I)$ with a state-conditioned choice of $w$ steers the policy's behavior while leaving the pretrained generative process intact~\citep{wagenmaker2025dsrl}.

\paragraph{Human-in-the-loop adaptation.}
We consider the standard interactive setting, where a human may intervene in behavior induced by a pretrained policy. At each step, the policy proposes an action; when the behavior is unsatisfactory, the expert takes over and provides a corrective action $a^*$ for the current observation $s$. This yields correction pairs $(s, a^*)$ on states the deployed policy actually visits, as in DAgger~\citep{ross2011dagger}. Our goal is to use these corrections to adapt the policy efficiently -- without modifying the base policy's weights or using a reward signal.

\section{Method}
\label{sec:method}
\methodname has two components: (1) \emph{action inversion} (\Cref{sec:method:inversion}) converts a corrective action into a target in the base policy's noise space and (2) a small \emph{noise policy} (\Cref{sec:method:flowdagger}) is trained on these targets to steer the frozen base policy at deployment. \Cref{sec:method:wam} extends action inversion to WAMs, whose action generation is entangled with future-state prediction. %

\subsection{Action Inversion}
\label{sec:method:inversion}
Latent-space adaptation methods typically rely on rewards or supervision already expressed in the policy's latent space. Human interventions, however, arrive as corrective actions. Action inversion converts these action-space corrections into noise-space supervision: given an observation $s$ and corrective action $a^*$, it recovers a noise vector $w^*$ such that the frozen generative policy reproduces the correction, $\pi_{\mathrm{gp}}(s, w^*) \approx a^*$; the pair $(s, w^*)$ is then a supervision target for the noise policy. 

The map $\pi_{\mathrm{gp}}(s, \cdot)$ composes $K$ Euler steps (\Cref{sec:background}) and has no closed-form inverse, so we invert it one step at a time. Each forward step $x_{k+1} = x_k + \Delta t\,v_\theta(x_k, t_k, s)$ rearranges to the implicit equation $x_k = x_{k+1} - \Delta t\,v_\theta(x_k, t_k, s)$, which we solve by fixed-point iteration:
\begin{equation}
\label{eq:fixed-point}
    x_k^{(m+1)} = x_{k+1} - \Delta t\,v_\theta\!\big(x_k^{(m)}, t_k, s\big), \qquad x_k^{(0)} = x_{k+1}.
\end{equation}
The right-hand side is a contraction in $x_k$ whenever $\Delta t\,L < 1$ ($L$ the Lipschitz constant of $v_\theta$ in its first argument), so the iteration converges geometrically and a few steps suffice; we use $M = 5$ (\Cref{app:inversion-iters}). Running backward from $x_K = a^*$ recovers $x_{K-1}, \dots, x_0$, and we take $w^* = x_0$.

Solving each step's implicit equation, rather than taking one explicit reverse pass, is what makes inversion reliable here. A single explicit reverse step evaluates the velocity at $x_{k+1}$ rather than $x_k$, an $O(\Delta t)$ discrepancy; this is negligible under image-diffusion schedules ($\Delta t \sim 10^{-3}$) but not for few-step action heads ($K=10$, $\Delta t = 0.1$). It is also more damaging here: robot actions are small, so an error negligible against image latents can be a large fraction of the action. The fixed-point solution of \Cref{eq:fixed-point} drives each step's residual to near zero independent of $\Delta t$; \Cref{app:inversion-procedure} compares against trajectory-level and optimization-based variants.

Inverting one correction costs $K\,M$ evaluations of $v_\theta$, requires no backpropagation through the base policy, and runs online as corrections are collected.

\subsection{Inverting World-Action Models}
\label{sec:method:wam}

The inversion of \Cref{sec:method:inversion} assumes a base policy whose action generation is a self-contained process in action space. WAMs break this assumption: they generate an action chunk jointly with future world states, as slices of a single latent video sequence~\citep{kim2026cosmospolicy, ye2026dreamzero}, so no action-only process exists to invert. \methodname nonetheless applies to this family, by inverting the joint generative process directly. We describe the procedure for Cosmos-Policy~\citep{kim2026cosmospolicy}, a WAM built on a pretrained video diffusion model.

\paragraph{Inverting the joint process.}
In Cosmos-Policy the generative variable is a joint latent tensor whose frames decode to the action chunk, predicted future states $s'$, and a value estimate (shape $C \times T \times H \times W = 16 \times 9 \times 28 \times 28$). The forward sampler decodes this whole tensor with a shared denoiser, so we invert the same joint process: the reverse-ODE recovers a joint-latent noise $w^*$ of the same shape. We build the inversion target from the base policy's own predicted clean latent $x_0^{\mathrm{base}}$, perturbing only the action frame: $x_0^* = x_0^{\mathrm{base}}$ with the expert-minus-base action delta added to the action frame -- a minimal-delta swap that stays on the base manifold -- while the state and value frames are held at the base policy's outputs. The reverse pass recovers a $w^*$ reproducing the full $x_0^*$, not just the action slice, and $(s, w^*)$ supervises the noise policy as before. Cosmos-Policy uses an EDM-style schedule rather than a flow-matching ODE; the per-step inversion of \Cref{sec:method:inversion} carries over with a small modification at the nonzero terminal noise level (\Cref{app:wam-inversion:edm}).

\subsection{\methodname: Human-in-the-Loop Adaptation}
\label{sec:method:flowdagger}

\paragraph{Noise policy.}
\methodname adapts the base policy through a small \emph{noise policy} $\pi^w$ -- an observation encoder followed by an MLP, with its own parameters $\phi$, separate from the frozen base policy. At deployment the base policy's standard noise draw $w \sim \mathcal{N}(0, I)$ is replaced by $\pi^w(s)$, which is deterministic at deployment and is the only trained component, so the executed action is
\begin{equation}
\label{eq:deploy}
    a = \pi_{\mathrm{gp}}\big(s,\, \pi^w(s)\big).
\end{equation}

\paragraph{Adaptation loop.}
\methodname collects corrections on-policy, in the DAgger fashion~\citep{ross2011dagger}. The current policy is deployed with a human operator in the loop; when the operator judges its behavior unsatisfactory, they intervene with a corrective action $a^*$, producing a correction pair $(s, a^*)$ on a state the deployed policy actually visited. Action inversion (\Cref{sec:method:inversion}) converts each correction into a noise-space target $(s, w^*)$, which is added to a training set $\mathcal{D}$. The noise policy is trained to regress these targets by minimizing
\begin{equation}
\label{eq:loss}
    \mathcal{L}(\phi) = \mathbb{E}_{(s, w^*)\sim\mathcal{D}}\big\|\pi^w(s) - w^*\big\|_2^2.
\end{equation}
Deployment, correction, inversion, and training are interleaved: $\mathcal{D}$ is aggregated across rounds, so later corrections target the states induced by the adapted policy. Corrections are sparse relative to the policy's own transitions, so regressing $\pi^w$ on corrections alone overfits the handful of intervened states. \methodname{} therefore structures $\mathcal{D}$ as two buffers: inverted corrections $(s, w^*)$ fill an intervention buffer, while the noise that produced each transition of the policy's \emph{successful} autonomous rollouts fills a second buffer. Every training batch draws in equal proportion from the two. The autonomous buffer keeps $\pi^w$ anchored to the base policy's behavior on states it already handles, while the fixed share keeps the sparse corrections prominent in every update.

\section{Experiments}
\label{sec:experiments}

Our evaluation is designed to answer four questions: (i) how sample- and compute-efficient is \methodname relative to existing adaptation methods; (ii) does noise-space adaptation transfer across base-policy families, including world-action models with no separable action head; (iii) does it preserve and leverage the base policy's behavioral prior better than fine-tuning; and (iv) does it deliver these gains on real hardware across a range of manipulation tasks.

\subsection{Setup}
\label{sec:experiments:setup}

\begin{figure}[t]
  \centering
  \resizebox{0.9\columnwidth}{!}{%
    \begin{minipage}{\columnwidth}
      \centering
      \rwimg{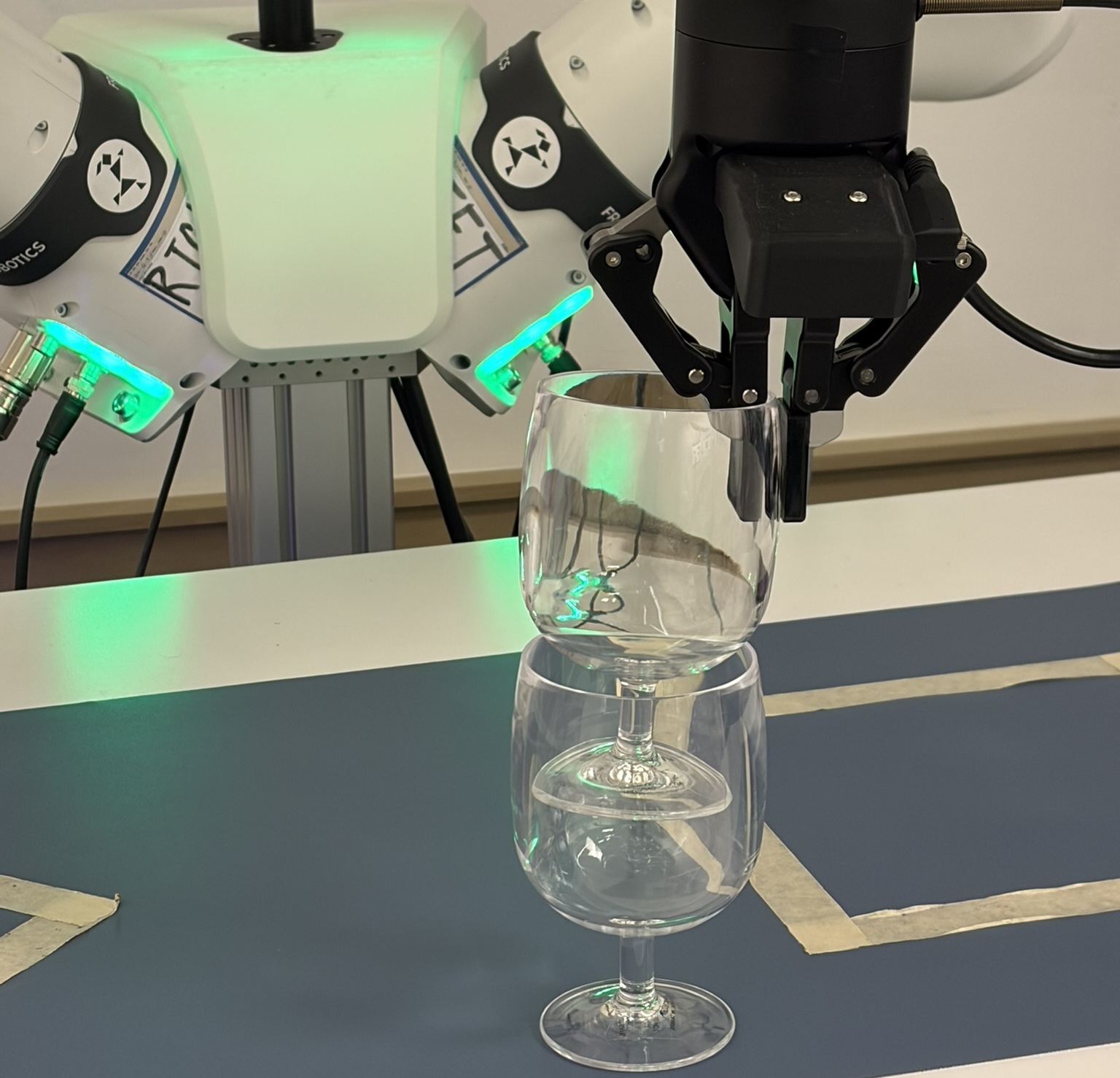}
      \rwimg{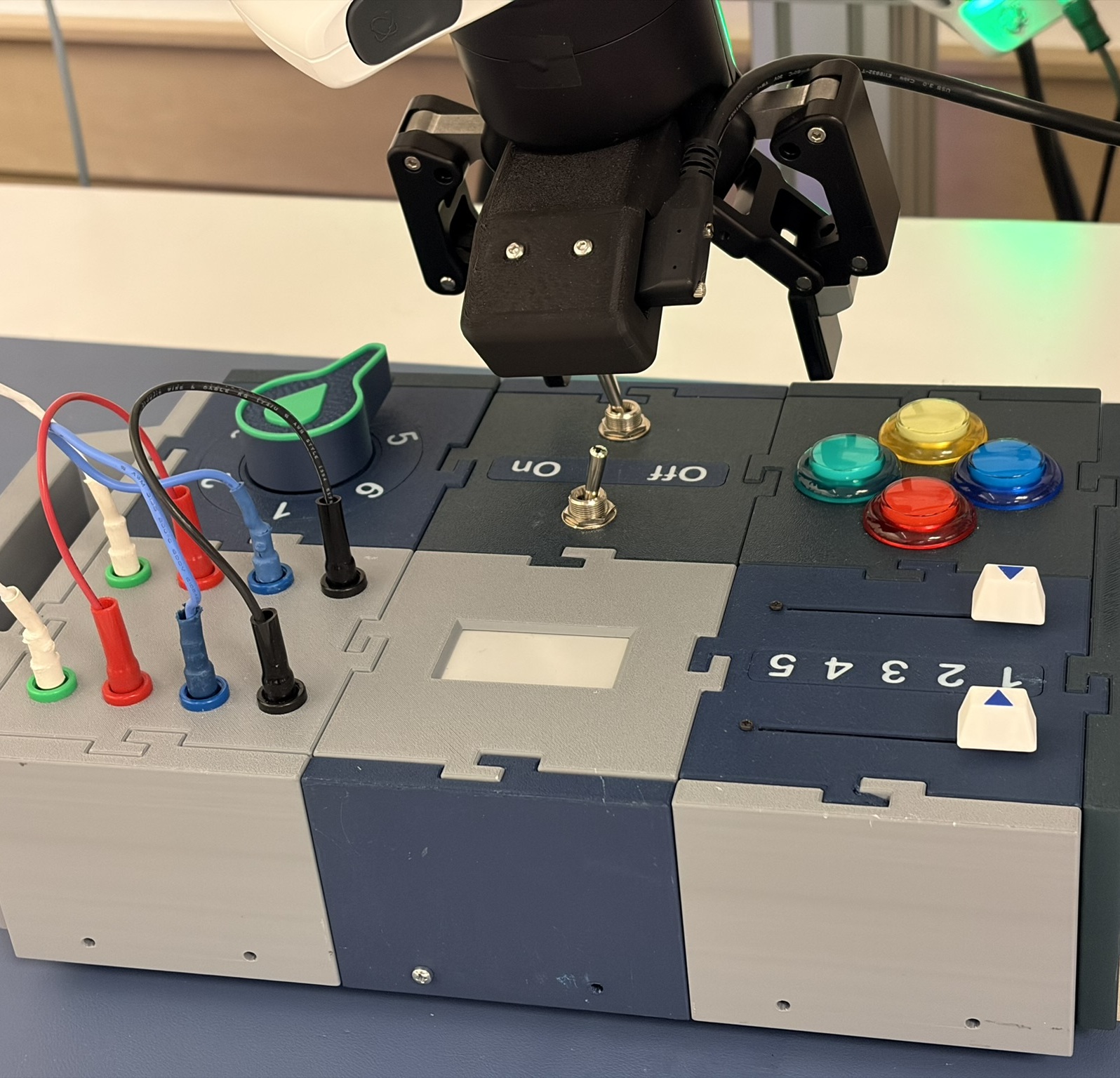}
      \rwimg{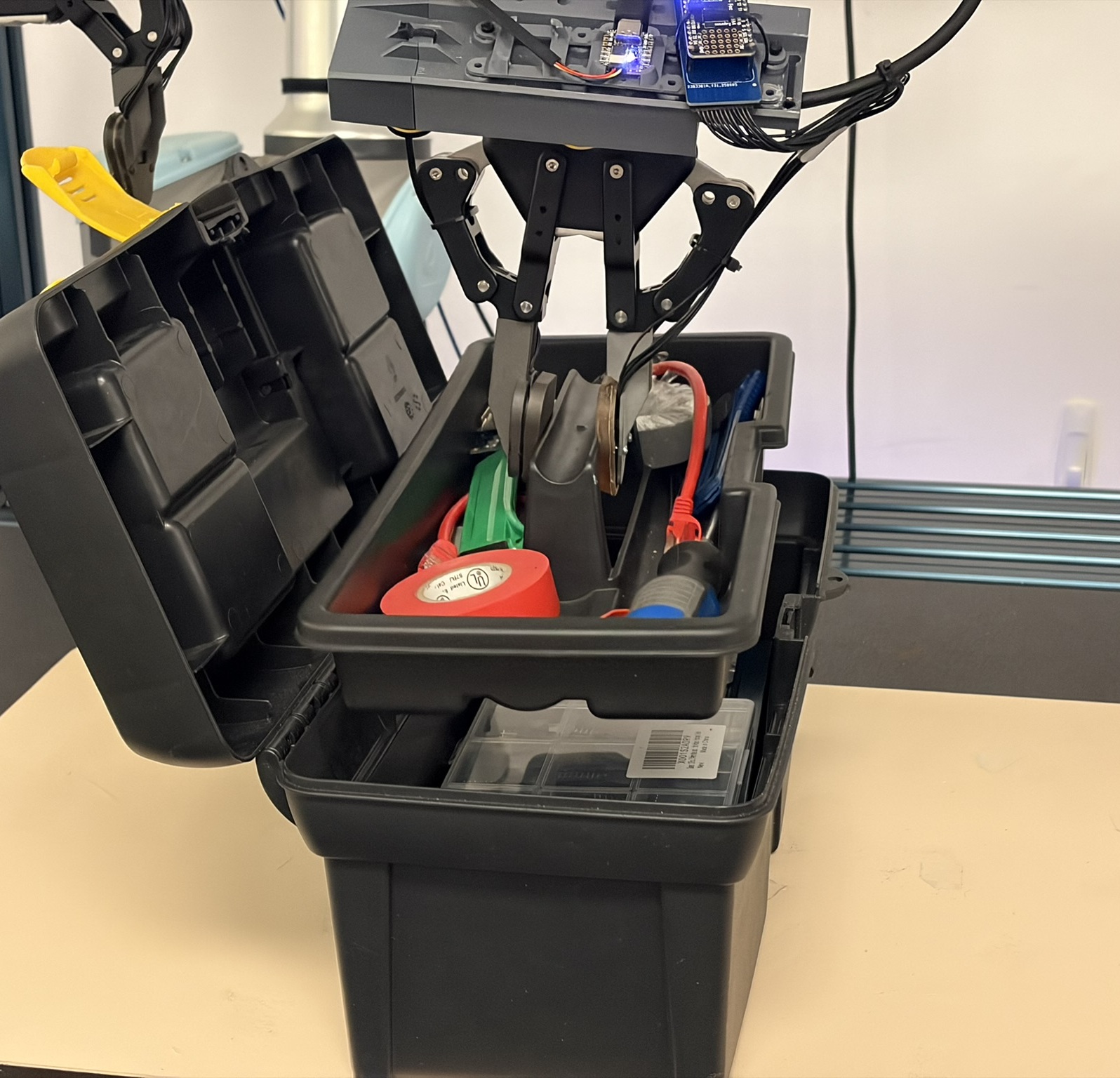}
      \rwimg{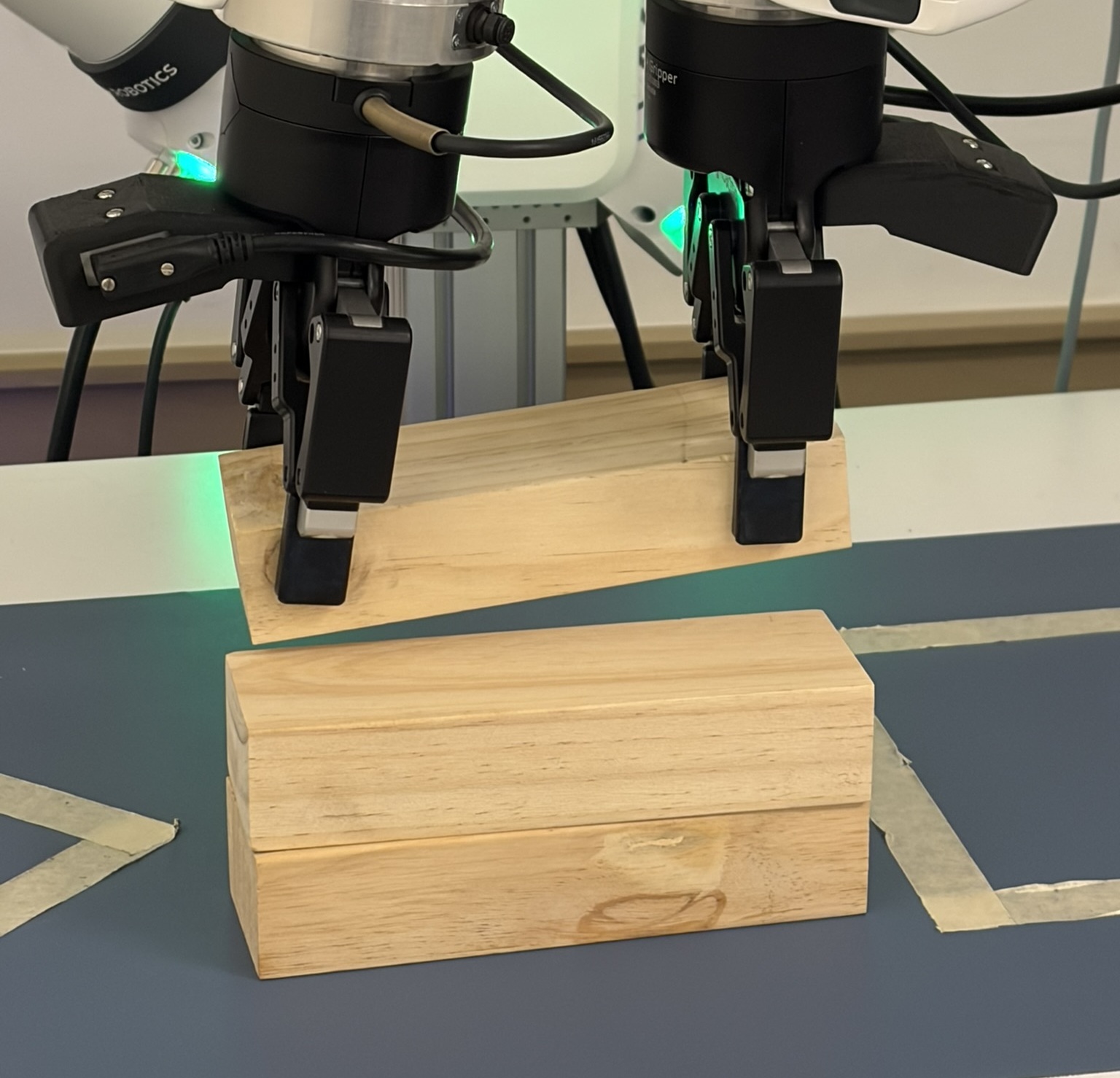}
      \rwimg{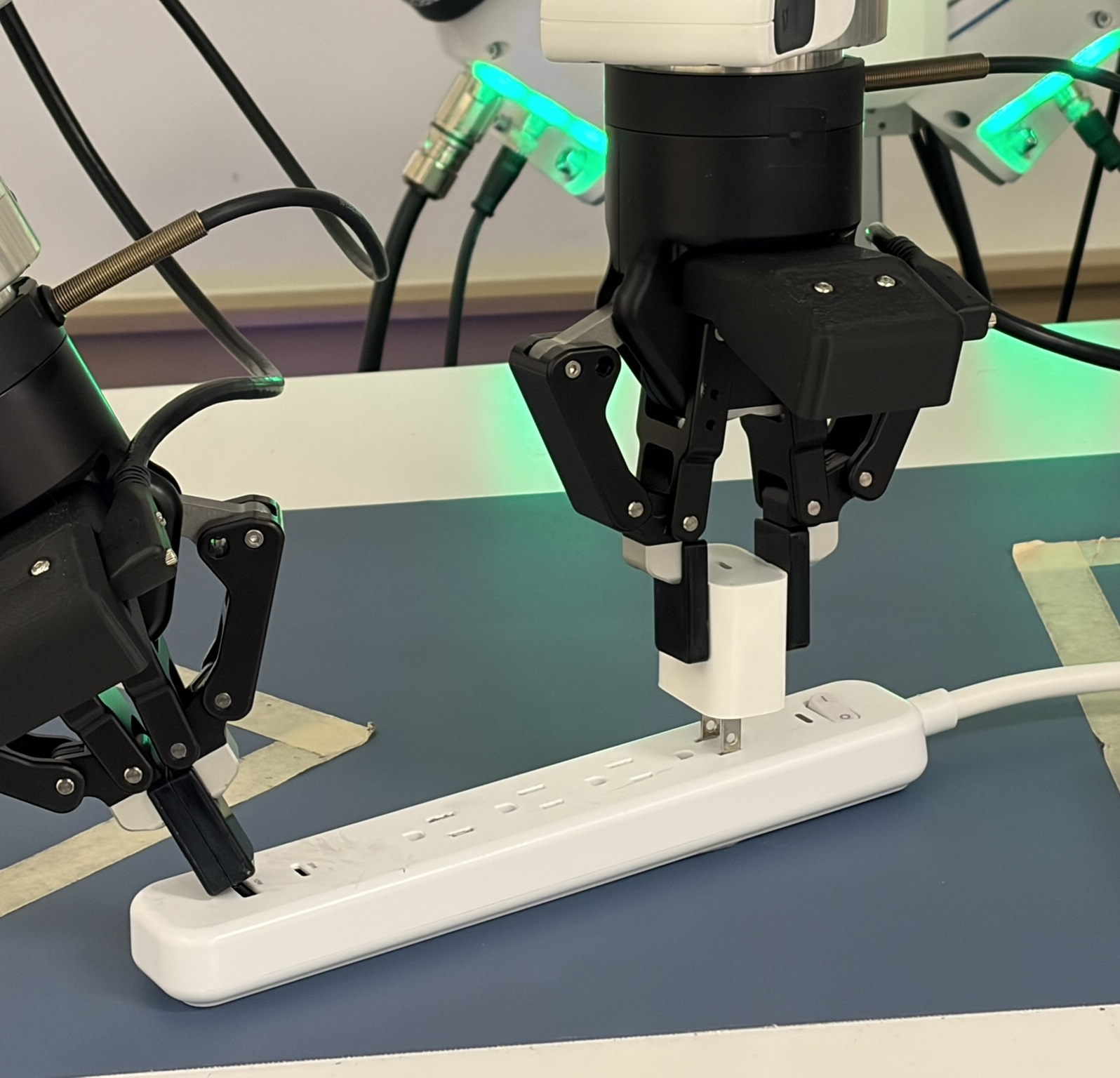}
    \end{minipage}%
  }
  \caption{Real-world evaluation tasks. \textit{Left to right:} Glassware Stacking, the BusyBox benchmark~\citep{fortier2026busybox} (Slider, Button, Wire Pull), Toolbox Packing, Jenga Stacking, and Plug Insertion.}
  \label{fig:real-world-setups}
\end{figure}

\paragraph{Tasks and platforms.}
We evaluate on a range of manipulation tasks in simulation and on real hardware, from basic pick-and-place to contact-rich and bimanual manipulation. In simulation we use MetaWorld~\citep{yu2019metaworld}, adapting on a diverse set of contact-rich tasks; the tasks evaluated per base policy are listed in \Cref{tab:method-comparison-pi05,tab:cross-base}. On real hardware we use two bimanual platforms, a \texttt{FR3 Duo} (two FR3 arms) and a \texttt{Dual UR5e} setup; depending on the task the policy controls a single arm or both. The unimanual tasks are Block Pick, Glassware Stacking, and three tasks from the BusyBox manipulation benchmark~\citep{fortier2026busybox} (Button Push, Slider, Wire Pull); the bimanual tasks are Jenga Stack, Toolbox Packing, and Plug Insertion. \Cref{fig:real-world-setups} shows representative shots of the tasks.

\paragraph{Base policies.}
We adapt a state-of-the-art representative of each major generative architecture family: $\pi_{0.5}$~\citep{intelligence2025pi05}, an action-head VLA, and Cosmos-Policy~\citep{kim2026cosmospolicy}, a WAM. $\pi_{0.5}$ uses the action-space inversion of \Cref{sec:method:inversion}; Cosmos-Policy exercises the joint world-action inversion of \Cref{sec:method:wam}. \Cref{app:additional-experiments} reports analogous results for a second VLA (Gr00t N1.7~\citep{nvidia2025groot}) and a vanilla diffusion policy~\citep{chi2023diffusion}, confirming the findings are not specific to these two base policies.

\paragraph{Baselines and protocol.}
We compare against baselines spanning the adaptation design space. The \emph{frozen base policy} fixes the starting performance. \emph{SFT} applies full-weight behavioral cloning to offline demonstrations, isolating the covariate-shift correction on-policy supervision provides. \emph{LoRA-DAgger} and \emph{Residual-DAgger} consume the same human-gated correction stream as \methodname but apply each correction in low-rank weights and an action-space residual, respectively, isolating the effect of adapting in noise space. \emph{DSRL}~\citep{wagenmaker2025dsrl} also adapts in noise space but learns from sparse reward and autonomous exploration rather than corrections, isolating the supervision signal. Unless noted, every method gets a matched budget of $N{=}50$ additional rollouts (on-policy with optional intervention for the correction-based methods, offline demonstrations for SFT, autonomous for DSRL), and we report mean success rate over $3$ seeds, $25$ rollouts each.

\subsection{\methodname{} adapts $\pi_{0.5}$ from a handful of corrections}
\label{sec:experiments:results}
\begin{figure}[t]
    \centering
    \includegraphics[width=0.9\columnwidth]{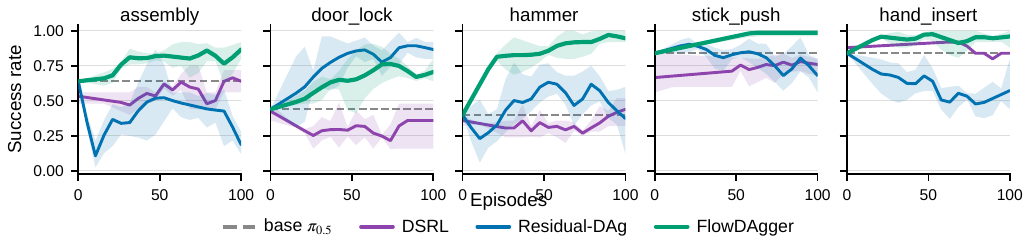}
    \caption{Success rate vs.\ adaptation rollouts for \methodname (green), Residual-DAgger (blue), and DSRL (purple) adapting frozen $\pi_{0.5}$ on five MetaWorld tasks. Dashed line: frozen base. Mean over $3$ seeds (shading bounds per-seed min/max), $25$ rollouts per point.}
    \label{fig:success-rate-pi05}
\end{figure}
\Cref{tab:method-comparison-pi05} compares \methodname against every baseline on twelve MetaWorld tasks with $\pi_{0.5}$ as the base policy. \methodname gives the largest mean improvement over the frozen base ($+0.25$) and wins on eight of the twelve tasks. The action- and weight-space DAgger baselines consume the same correction stream but gain far less, so the advantage comes from \emph{where} the correction is applied, not from the corrections alone. DSRL barely moves the base: a sparse reward and autonomous exploration cannot discover in $50$ rollouts the corrective behavior a handful of interventions supply directly. \Cref{fig:success-rate-pi05} shows the same over training: \methodname climbs fastest and highest and rises smoothly on all five tasks, while Residual-DAgger learns erratically, oscillating on Hammer and climbing then collapsing below the base on Assembly and Hand Insert.

An unconstrained action-space correction is well-conditioned only when a small, consistent offset suffices; Door Lock is the one such case here, where Residual-DAgger climbs smoothly and edges ahead. Steering in noise space stays on the base manifold, so adaptation remains stable whatever the correction's magnitude.

\begin{table}[t]
\centering
\caption{Method comparison on MetaWorld with $\pi_{0.5}$ as the base policy. The bottom rows give mean success rate and improvement over the frozen base. Best non-base method per row in \textbf{bold}.}
\label{tab:method-comparison-pi05}
\setlength{\tabcolsep}{4pt}
\small
\begin{tabular}{lrrrrrr}
\toprule
Task & Base & SFT & LoRA-DAg.\ & Res-DAg.\ & DSRL & \methodname \\
\midrule
Assembly         & 0.64 & 0.85 & 0.81 & 0.53 & 0.64 & \textbf{0.89} \\
Bin Picking      & 0.56 & \textbf{0.76} & 0.68 & 0.63 & \textbf{0.76} & 0.69 \\
Box Close        & 0.36 & \textbf{0.70} & 0.52 & 0.69 & 0.48 & 0.59 \\
Coffee Pull      & 0.84 & 0.96 & 0.68 & 0.95 & 0.84 & \textbf{1.00} \\
Dial Turn        & 0.20 & 0.64 & \textbf{0.88} & 0.59 & 0.43 & 0.75 \\
Door Lock        & 0.44 & 0.48 & 0.76 & \textbf{0.85} & 0.28 & 0.75 \\
Hammer           & 0.40 & 0.80 & 0.68 & 0.56 & 0.27 & \textbf{0.84} \\
Hand Insert      & 0.84 & 0.88 & 0.72 & 0.71 & 0.92 & \textbf{0.99} \\
Lever Pull       & 0.28 & 0.44 & 0.52 & 0.37 & 0.35 & \textbf{0.61} \\
Pick Place       & 0.76 & 0.80 & 0.68 & 0.71 & 0.60 & \textbf{0.85} \\
Soccer           & 0.24 & 0.36 & 0.32 & 0.28 & 0.33 & \textbf{0.44} \\
Stick Push       & 0.84 & 0.84 & 0.92 & 0.84 & 0.73 & \textbf{1.00} \\
\midrule

Mean SR          & 0.53 & 0.71 & 0.68 & 0.64 & 0.55 & \textbf{0.78} \\
$\Delta$ vs.\ Base & --   & $+0.18$ & $+0.15$ & $+0.11$ & $+0.02$ & $\mathbf{+0.25}$ \\
\bottomrule
\end{tabular}
\end{table}

\begin{wrapfigure}[11]{r}{0.34\columnwidth}
  \centering
  \vspace{-2.7em}
  \includegraphics[width=\linewidth]{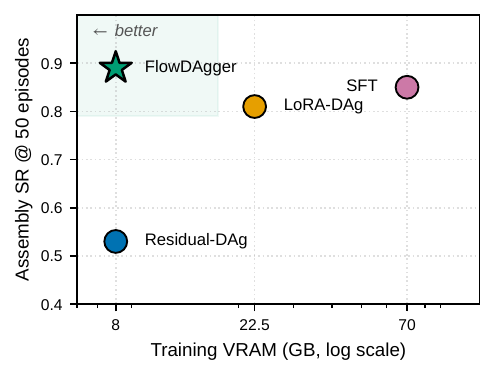}
  \caption{Assembly success rate at $N{=}50$ vs.\ peak training VRAM (upper-left is better).}
  \label{fig:compute-efficiency}
\end{wrapfigure}

\paragraph{Compute efficiency.}
Because \methodname trains only a small noise policy, adaptation fits in the same $\sim\!8$~GB that deploying $\pi_{0.5}$ already requires. It can therefore run on a single consumer GPU. \Cref{fig:compute-efficiency} plots success rate against this training footprint, where \methodname sits alone in the upper-left -- as accurate as any baseline at a fraction of the memory.

\subsection{Adaptation transfers across base-policy families}

The same noise-space adaptation procedure works across generative policy families. \Cref{tab:cross-base} applies \methodname{} to $\pi_{0.5}$, an action-head VLA with action-space inversion, and to Cosmos-Policy, a WAM with joint world-action inversion. The gains are comparable on both, even though Cosmos-Policy produces actions only as slices of a joint world-action latent rather than from a separable action head. This suggests that \methodname{} does not depend on an action-only generative process. \Cref{app:additional-experiments} extends the result to Gr00t N1.7 and a diffusion policy.

\begin{table}[t]
\centering
\caption{\methodname across base-policy families, on a shared set of MetaWorld tasks. $\Delta$ is the improvement over each frozen base.}
\label{tab:cross-base}
\setlength{\tabcolsep}{5pt}
\small
\begin{tabular}{lrrrcrrr}
\toprule
& \multicolumn{3}{c}{$\pi_{0.5}$} & & \multicolumn{3}{c}{Cosmos-Policy} \\
\cmidrule{2-4} \cmidrule{6-8}
Task & Base & \methodname & $\Delta$ & & Base & \methodname & $\Delta$ \\
\midrule
Assembly       & 0.64 & 0.89 & $+0.25$ & & 0.52 & 0.92 & $+0.40$ \\
Bin Picking    & 0.56 & 0.69 & $+0.13$ & & 0.28 & 0.44 & $+0.16$ \\
Box Close      & 0.36 & 0.59 & $+0.23$ & & 0.36 & 0.64 & $+0.28$ \\
Dial Turn      & 0.20 & 0.75 & $+0.55$ & & 0.44 & 0.68 & $+0.24$ \\
Hand Insert    & 0.84 & 0.99 & $+0.15$ & & 0.76 & 0.88 & $+0.12$ \\
Lever Pull     & 0.28 & 0.61 & $+0.33$ & & 0.56 & 0.64 & $+0.08$ \\
Stick Push     & 0.84 & 1.00 & $+0.16$ & & 0.76 & 0.96 & $+0.20$ \\
\midrule
Mean           & 0.53 & 0.79 & $+0.26$ & & 0.53 & 0.74 & $+0.21$ \\
\bottomrule
\end{tabular}
\end{table}

\subsection{Noise-space adaptation preserves the base prior}

Because \methodname{} steers the frozen policy through its noise input, adaptation should change the task behavior without rewriting unrelated skills. We test this by adapting each method on Hammer ($50$ episodes; $50$ demos for SFT at a matched gradient-step budget) and evaluating on five held-out tasks that the base policy nearly saturates. \methodname{} preserves the held-out tasks best while also achieving the largest in-distribution gain. Weight-space baselines buy their Hammer improvement by collapsing held-out performance, while Residual-DAgger preserves the prior only because it barely adapts. Only \methodname{} does both.

\begin{table}[t]
\centering
\caption{Prior preservation on $\pi_{0.5}$: \methodname{} holds the held-out tasks closest to the frozen base while gaining the most in-distribution. \textbf{Bold}: held-out scores within base noise or above.}
\label{tab:prior-preservation}
\setlength{\tabcolsep}{4pt}
\small
\begin{tabular}{lcccccccc}
\toprule
& \multicolumn{1}{c}{\textbf{In-dist}} & \multicolumn{6}{c}{\textbf{Held-out (saturated) tasks}} & \\
\cmidrule(lr){2-2} \cmidrule(lr){3-8}
Method & Hammer & Door & Drawer & Faucet & Plate & Push & Mean & $\Delta$ vs base \\
\midrule
Base $\pi_{0.5}$ (no adapt)        & $0.40$ & $\mathbf{1.00}$ & $\mathbf{1.00}$ & $\mathbf{0.96}$ & $\mathbf{0.88}$ & $\mathbf{0.96}$ & $\mathbf{0.96}$ & --- \\
\midrule
\textbf{\methodname{}}             & $\mathbf{0.84}$ & $0.88$          & $\mathbf{1.00}$ & $\mathbf{0.96}$ & $\mathbf{1.00}$ & $0.56$          & $0.88$          & $-0.08$ \\
Residual-DAg                       & $0.56$ & $0.88$          & $0.68$          & $0.80$          & $\mathbf{1.00}$ & $0.08$          & $0.69$          & $-0.27$ \\
LoRA-DAg                           & $0.68$ & $0.80$          & $0.00$          & $0.36$          & $0.00$          & $0.36$          & $0.30$          & $-0.66$ \\
SFT ($50$ demos, openpi finetune)  & $0.80$ & $0.12$          & $0.00$          & $0.00$          & $0.00$          & $0.00$          & $0.02$          & $-0.94$ \\
\bottomrule
\end{tabular}
\end{table}

\

\subsection{Real-world adaptation}

\begin{wraptable}{r}{0.58\columnwidth}
\vspace{-3em}
\setlength{\tabcolsep}{3pt}
\footnotesize
\centering
\caption{Real-hardware success over $30$ rollouts for the frozen base, the \textsc{Sft}\ baseline, \methodname, and the number of additional episodes. Best non-base per row in \textbf{bold}.}
\label{tab:real-world}
\begin{tabular}{lrrrr}
\toprule
Task & Base & SFT & \methodname\ ($\Delta$) & \# Eps. \\
\midrule
Block Pick           & 0.73 & 0.80 & \textbf{0.90} ($+0.17$) & 5 \\
Glassware Stacking   & 0.26 & 0.53 & \textbf{0.76} ($+0.50$) & 5 \\
Button Push          & 0.60 & \textbf{0.73} & \textbf{0.73} ($+0.13$) & 10 \\
Slider               & 0.36 & 0.43 & \textbf{0.66} ($+0.30$) & 5 \\
Wire Pull            & 0.40 & 0.53 & \textbf{0.70} ($+0.30$) & 10 \\
Jenga Stacking       & 0.76 & \textbf{0.90} & 0.86 ($+0.10$) & 5 \\
Toolbox Packing      & 0.13 & 0.63 & \textbf{0.80} ($+0.67$) & 10 \\
Plug Insertion       & 0.60 & 0.66 & \textbf{0.72} ($+0.12$) & 20 \\
\bottomrule
\end{tabular}
\end{wraptable}

We evaluate on two real robot platforms (an \texttt{FR3 Duo} and a \texttt{Dual UR5e}) across eight tasks spanning single-arm, bimanual, and contact-rich manipulation. With only five to twenty intervention episodes, \methodname{} substantially improves the frozen base on every task. On Toolbox Packing, for example, ten corrections raise success from $13\%$ to $80\%$, turning a largely unusable policy into a reliable one without modifying the foundation model (\Cref{tab:real-world}). \methodname{} also outperforms SFT, a full fine-tuning baseline, on every task except Jenga Stacking, where SFT edges ahead.

\vspace{-0.5em}
\section{Limitations}
\label{sec:limitations}
\vspace{-0.5em}
Like other latent-space adaptation methods, \methodname adapts behavior by steering a frozen generative policy through its latent variables rather than modifying the policy itself. Its adaptation capacity is therefore bounded by the support of the pretrained model: if a desired behavior lies far outside the policy's action manifold, inversion can only recover the closest behavior representable by the base policy. In such cases, additional data or weight-space adaptation may be needed.

Adaptation also depends on the quality and coverage of human interventions. As with other DAgger-style methods, \methodname can only improve behavior in regions of the state space where corrective supervision is provided, and systematic biases or inconsistencies in those corrections may be reflected in the adapted policy. Finally, inversion accuracy can degrade for highly multimodal action distributions or poorly conditioned generative dynamics, reducing the fidelity with which expert corrections are represented in latent space.

\vspace{-0.5em}
\section{Conclusion}
\label{sec:conclusion}
\vspace{-0.5em}
We introduced \methodname, a framework for adapting pretrained generative robot policies from human interventions in latent space. Action inversion maps corrective actions to the latent variables that would have produced them, providing supervision for a lightweight noise policy while keeping the foundation model frozen. Across simulation, real robots, action-head VLAs, diffusion policies, and world-action models, \methodname improves from only a handful of interventions while preserving pretrained capabilities.

More broadly, \methodname treats interaction as a way to supervise the latent decisions of a frozen generative policy. Instead of modifying a foundation model's parameters to absorb new experience, we recover the latent decisions that would have produced expert behavior and learn to reproduce them directly. This perspective decouples adaptation from model scale, preserves pretrained capabilities, and gives a practical mechanism for improving robot foundation models through interaction while keeping them stable, reusable, and computationally efficient.

\clearpage

\bibliography{references}  %

\newpage

\appendix
\crefalias{section}{appendix}
\crefalias{subsection}{appendix}
\section{World-Action Model Inversion Details}
\label{app:wam-inversion}

This appendix expands the world-action model inverter sketched in \Cref{sec:method:wam}. \Cref{app:wam-inversion:edm} gives the full EDM-schedule inversion procedure used for Cosmos-Policy; \Cref{app:wam-inversion:postmortem} documents an action-frame-only variant that an earlier version of the inverter optimized and that the joint-process inversion of \Cref{sec:method:wam} replaces; \Cref{app:wam-inversion:steering} describes how the noise policy is parameterized for the resulting high-dimensional joint noise.

\subsection{Inverting the EDM diffusion schedule}
\label{app:wam-inversion:edm}

Cosmos-Policy generates with an EDM-style diffusion process rather than a flow-matching ODE, which changes the recurrence inversion must undo. The noise schedule does not terminate at zero noise: it runs from a large $\sigma_{\max}$ down to a nonzero $\sigma_{\min}$ (here $\sigma_{\max} = 80$, $\sigma_{\min} = 4$), and the forward sampler we invert is a 1-Euler integrator over $\sigma_{\max} \!\to\! \sigma_{\min}$ followed by a single terminal denoise $D(\cdot, \sigma_{\min})$. The 1-Euler integration is inverted with the same per-step fixed-point iteration as in \Cref{sec:method:inversion}, applied to the reverse of the EDM update; because each step depends only on the current latent and noise level, the reversal is memoryless and gradient-free. The terminal denoise at $\sigma_{\min}$ is a strong transform -- the schedule does not bottom out near identity -- so a single explicit reverse step cannot invert it. We recover its input with a short local Adam solve anchored at the base policy's own pre-terminal state, backpropagating through one denoiser call per Adam step. This terminal solve is the only part of the inversion that backpropagates through the base model; like the rest of the inverter it requires no backprop through the chained sampler and runs online as corrections are collected.

\subsection{Why action-frame-only inversion is insufficient}
\label{app:wam-inversion:postmortem}

An earlier variant of the inverter optimized only the action-frame noise (112 dimensions in our configuration) with Adam, held the non-action frames at a fixed reference, and minimized the action-reconstruction loss through the chained denoiser calls of the forward sampler. This consistently floored at a non-trivial reconstruction error -- about $0.04$ even when inverting the base policy's own actions and $0.076$ on real expert actions -- and grew worse with more denoising steps, because the chained-denoiser objective is sharply non-convex in the action-frame noise alone. Inverting the joint process instead replaces that hard optimization with the exact per-step fixed-point inverse described in \Cref{sec:method:wam} and a single small autograd-based solve at $\sigma_{\min}$.

\subsection{Parameterizing the noise policy for world-action models}
\label{app:wam-inversion:steering}

For action-head base policies the noise is action-chunk shaped and low-dimensional enough that the noise policy $\pi^w$ regresses it directly. Because $\pi^w$ is trained by supervised regression on inverted-noise targets rather than by reinforcement learning, it tolerates this dimensionality without the reductions that noise-space RL requires: DSRL~\citep{wagenmaker2025dsrl}, for instance, collapses $\pi_0$'s $1{,}600$-dimensional chunk noise to a single per-step vector repeated across the chunk to keep SAC stable.

World-action models are different. The joint noise returned by the inverter is of order $10^5$ dimensions, too high for $\pi^w$ to regress directly even under supervision. We compared two options for $\pi^w$'s output. \emph{Full}: regress the entire joint latent -- maximum capacity but no prior. \emph{Basis}: regress $k{=}64$ PCA coefficients fit on inverted-noise targets collected during a brief base-policy warmup, anchoring the output to the manifold of successful base-policy noises. Both worked; the basis parameterization was somewhat more stable and is the default in our Cosmos-Policy experiments. Either choice is downstream of inversion, which always operates on the full joint process.

\section{Additional Experiments}
\label{app:additional-experiments}

This appendix reports supplementary experimental results. \Cref{app:inversion-procedure} gives the full inversion-procedure comparison and \Cref{app:inversion-iters} the iteration-count sweeps. \Cref{app:additional-experiments:groot,app:additional-experiments:dp} replicate the main-paper experiments with two additional pretrained generative policies as the base model -- Gr00t N1.7~\citep{nvidia2025groot} (a second action-head VLA) and a vanilla diffusion policy~\citep{chi2023diffusion} -- to confirm that the findings are not specific to $\pi_{0.5}$ and Cosmos-Policy.

\subsection{Inversion procedure comparison}
\label{app:inversion-procedure}

We compare the per-step fixed-point inversion of \Cref{sec:method:inversion} against (i) a single explicit Euler reverse pass, (ii) trajectory-level fixed-point inversion, which corrects the global trajectory-reconstruction residual~\citep{rout2024rfinversion}, and (iii) optimization-based inversion (Adam on an action-reconstruction loss backpropagated through the frozen decoder). All variants invert the same scripted-expert chunks on the assembly task and reconstruct through the frozen $\pi_{0.5}$ decoder with the standard $K{=}10$-step Euler schedule (\Cref{tab:inversion-procedure}).

Per-step fixed-point inversion is roughly an order of magnitude more accurate than every trajectory-level alternative -- at $M{=}5$ its Action MSE is $\sim\!20\times$ lower than Euler reverse and $\sim\!14\times$ lower than trajectory-level FP -- and $\sim\!3\times$ faster per chunk. Cranking the baselines does not close the gap: trajectory FP saturates by $k{=}5$ and a 20-step Adam refinement only matches a single Euler pass (\Cref{app:inversion-iters}). Both correct a global residual they cannot attribute to any single step, and at $\Delta t{=}0.1$ the per-step linearization errors are too large to remove this way. This is consistent with trajectory-level inversion results for image diffusion and rectified flow~\citep{mokady2023nulltext,rout2024rfinversion}, established under the far smaller per-step errors of $50$--$1000$-step schedules; few-step flow-matching action heads sit in a different regime, and the per-step solve is what restores accuracy there.

These reconstruction gaps carry downstream. Training the noise policy on each inverter's targets, per-step FP reaches $0.87$/$0.96$ SR on assembly/hammer versus $0.59$/$0.71$ for Euler reverse, with the trajectory-level and Adam variants clustered between (their order swaps across tasks within seed noise). Inaccurate inversion is not merely slower: Euler-reverse supervision pushes the noise policy \emph{below} the frozen base on assembly ($0.59$ vs $0.64$), since targets noisier than $\sim\!0.03$ Action MSE corrupt rather than steer the policy. Accurate inversion therefore sets a strictly higher ceiling, not just a faster path to it.

\begin{table}[t]
\centering
\caption{Inversion-procedure comparison on $\pi_{0.5}$ + MetaWorld (assembly), each row inverting the same $29$ scripted-expert chunks. Action MSE is in the 4-D env action space (per-dim range $[-1,1]$); wall-clock is the median per-chunk time on an RTX~5090 after warmup. The SR columns report downstream task success after training the noise policy on each inverter's targets ($3$ seeds, $25$ rollouts). Per-step FP at $M{=}5$ is the \methodname default; $M$/$k$ sweeps are in \Cref{app:inversion-iters}.}
\label{tab:inversion-procedure}
\setlength{\tabcolsep}{5pt}
\small
\begin{tabular}{lrrrr}
\toprule
Method & Action MSE & Time (ms) & SR (Assembly) & SR (Hammer) \\
\midrule
Euler reverse                                  & 0.0329 & 315 & 0.59 & 0.71 \\
Optimization (Adam, 20 steps, lr $10^{-2}$)    & 0.0275 & 1350 & 0.79 & 0.75 \\
Trajectory FP ($k{=}5$)                        & 0.0228 & 1662 & 0.77 & 0.68 \\
\textbf{Per-step FP ($M{=}5$)} (default)       & $\mathbf{0.00168}$ & $\mathbf{456}$ & $\mathbf{0.87}$ & $\mathbf{0.96}$ \\
\bottomrule
\end{tabular}
\end{table}

\subsection{Inversion iteration counts}
\label{app:inversion-iters}

\Cref{tab:inversion-iters} sweeps the inner iteration count $M$ of per-step fixed-point inversion (\Cref{eq:fixed-point}) and the outer iteration count $k$ of trajectory-level fixed-point inversion, with all other settings matching \Cref{tab:inversion-procedure}. The two sweeps are not directly comparable on a per-iteration basis because $k$ and $M$ parametrize different objects: each $k$ iteration is a full forward-plus-reverse pass over the $K{=}10$-step trajectory ($20$ model evaluations), while each $M$ iteration is a single velocity evaluation at one of the $K$ reverse steps. The natural matched-compute pairing is therefore $k{=}5$ ($110$ model evals) against $M{=}10$ ($110$ model evals); even at matched compute, per-step FP yields $\sim\!15\times$ lower Action MSE than trajectory-level FP, and runs $\sim\!2.7\times$ faster.

For per-step FP, Action MSE drops $2.4\times$ from $M{=}3$ to $M{=}5$, then only $1.1\times$ further at $M{=}10$ while wall-clock rises by $1.3\times$; the per-step contraction has effectively converged by $M{=}5$, which we use as the default. For trajectory-level FP, raising $k$ from $5$ to $20$ produces no measurable improvement -- the algorithm has already extracted all the information it can from the global residual, and the remaining error is the irreducible per-step linearization error that the trajectory-level formulation cannot address.

\begin{table}[t]
\centering
\caption{Iteration-count sweeps for the two fixed-point inversion variants. Action MSE and wall-clock are measured on the assembly task with $\pi_{0.5}$ as the base policy, under the same setup as \Cref{tab:inversion-procedure}.}
\label{tab:inversion-iters}
\setlength{\tabcolsep}{6pt}
\small
\begin{tabular}{lrrr}
\toprule
Method & Action MSE & Model evals & Time (ms) \\
\midrule
Trajectory FP ($k{=}5$)    & 0.0228   & 110 & 1662 \\
Trajectory FP ($k{=}20$)   & 0.0237   & 410 & 5657 \\
\midrule
Per-step FP ($M{=}3$)      & 0.00397  & 40  & 405 \\
Per-step FP ($M{=}5$)      & 0.00168  & 60  & 456 \\
Per-step FP ($M{=}10$)     & 0.00147  & 110 & 614 \\
\bottomrule
\end{tabular}
\end{table}

\subsection{Gr00t N1.7}
\label{app:additional-experiments:groot}

We replace $\pi_{0.5}$ with Gr00t N1.7~\citep{nvidia2025groot}, a second action-head VLA built on a Qwen3-VL backbone with a DiT flow-matching head. We use the LIBERO-10 checkpoint released by the authors and evaluate on LIBERO-90 task~57 (\textit{pick up the cream cheese and put it in the tray}), which is held out from the checkpoint's training set and zero-shots at 60\% success. Both methods adapt the same frozen base policy from 50 episodes of additional online experience; \methodname uses scripted-expert interventions, while DSRL uses task reward only. To make SAC tractable in Gr00t's $40 \times 132 = 5{,}280$-dimensional noise space, we project noise onto a $K{=}10$ DCT-II polynomial basis ($1{,}320$-dimensional effective action space).

\begin{figure}[t]
    \centering
    \includegraphics[width=0.6\columnwidth]{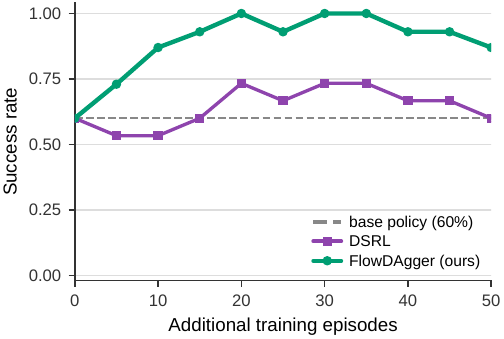}
    \caption{Success rate vs.\ additional training episodes for \methodname (green) and DSRL (purple) adapting frozen Gr00t N1.7 on LIBERO-90 task~57. Dashed line is the frozen base policy's zero-shot success rate. Single seed; each evaluation point averages $15$ rollouts. \methodname matches the base policy after $5$ episodes and saturates near $100\%$ by episode $20$; DSRL improves above base but plateaus around $70\%$.}
    \label{fig:appendix-groot-libero}
\end{figure}

\Cref{fig:appendix-groot-libero} shows that the qualitative finding from the main paper transfers to Gr00t: \methodname's noise-space supervision rapidly drives the success rate to ceiling, while DSRL is restricted to the much smaller signal of task reward and converges to a lower plateau roughly $25$ percentage points below \methodname's peak.

\subsection{Diffusion Policy}
\label{app:additional-experiments:dp}

We additionally evaluate on a vanilla DDPM-based diffusion policy~\citep{chi2023diffusion}, adapting it on the robomimic LIFT task~\citep{mandlekar2021robomimic} from low-dimensional state observations. The base policy zero-shots at $\approx 48\%$, and both methods are given $50$ additional online episodes.

\begin{figure}[t]
    \centering
    \includegraphics[width=0.6\columnwidth]{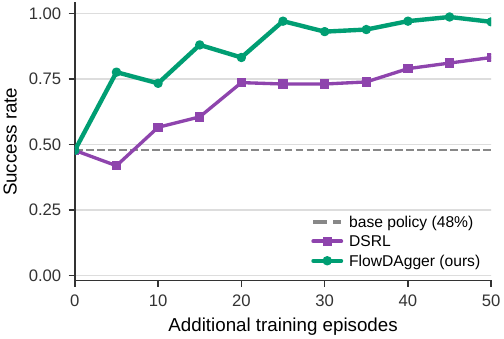}
    \caption{Success rate vs.\ additional training episodes for \methodname (green) and DSRL (purple) adapting a frozen diffusion policy on robomimic LIFT (state-only). Dashed line is the frozen base policy's success rate. Single seed; each evaluation point averages $15$ rollouts. \methodname reaches $\sim 97\%$ by episode $25$; DSRL also climbs but plateaus around $80\%$.}
    \label{fig:appendix-dp-lift}
\end{figure}

\Cref{fig:appendix-dp-lift} confirms that the picture extends to a diffusion-based generative policy on a different embodiment: \methodname is again the faster learner and reaches a higher asymptote, though the gap to DSRL is smaller than on Gr00t. We attribute this to two factors: the much lower-dimensional action and noise space of this base model makes DSRL's task-reward-only search more tractable, and LIFT's dense progress structure makes the reward signal informative even without expert supervision.

\end{document}